# Reformulation is All You Need: Addressing Malicious Textual Features in DNNs

Yi Jiang[1,2] Oubo Ma[1] Yong Yang[1] Tong Zhang[1] Shouling Ji[1]

[1]College of Computer Science and Technology, Zhejiang University, Hangzhou, 310007, China

[2]Guizhou Medical University, Guiyang, 550004, China

**Abstract:** Human language encompasses a wide range of intricate and diverse implicit features, which attackers can exploit to launch adversarial or backdoor attacks, compromising DNN models for NLP tasks. Existing model-oriented defenses often require substantial computational resources as the model size increases whereas sample-oriented defenses typically focus on specific attack vectors or schemes, rendering them vulnerable to adaptive attacks. We observe that the root cause of both adversarial and backdoor attacks lies in the encoding process of DNN models, where subtle textual features, negligible for human comprehension, are erroneously assigned significant weights by less robust or trojaned models. On this basis, we propose a unified and adaptive defense framework that is effective against both adversarial and backdoor attacks. Our approach leverages reformulation modules to address potential malicious features in textual inputs while preserving the original semantic integrity. Extensive experiments demonstrate that our framework outperforms existing sample-oriented defense baselines across a diverse range of malicious textual features.

**Keywords:** Adversarial attack, Backdoor attack, Defense, Textual features, Reformulation.

## 1 Introduction

Deep neural networks (DNNs) initially gained significant attention in the field of computer vision[1-2] and have since achieved remarkable breakthroughs in natural language processing (NLP)[3-5]. Today's generative large-scale models are even advancing the field toward artificial general intelligence (AGI)[6-7]. Despite these successes, the reliability of DNNs remains questionable because of their lack of interpretability[8-9], like a sword of Damocles hanging over the deployment of these models, which hinders the deployment of DNNs in safety-critical applications, where trust and predictability are paramount. The intrinsic properties of neural networks make them vulnerable to 2 notorious attack vectors: adversarial example attacks[10-11] and backdoor attacks[12-15]. Subtle perturbations added to inputs or training data can lead to unintended model behaviors, posing significant risks in practical applications.

With the growing adoption of NLP applications in DNN models, textual features such as typos[16], rare words[17], unique styles[15], and syntactic structures[14] can be exploited by attackers. While often overlooked in everyday use, these subtle characteristics enable adversarial example generation or data poisoning, subtly manipulating model behavior. By injecting rare features or modifying labels, attackers can craft adversarial samples to mislead clean models or implant backdoors to create trojaned models.

To mitigate these threats, extensive research in both industry and academia has led to the development of diverse defense strategies. Against adversarial attacks, approaches such as adversarial training enhance model robustness, perturbation control mitigates input-level threats, and certification-based methods provide risk bounds [18-20]. For backdoor attacks, defenses include model pruning, defensive distillation, and training adjustments to remove hidden backdoors, along with reverse engineering triggers and neuron analysis to detect compromised models [21-22].

**Motivation.** The existing defense mechanisms against adversarial threats can be categorized into two broad types. The first type[23-25] can be considered model-oriented, focusing on enhancing model robustness and eliminating or detecting backdoors at the model level. As the number of model parameters increases, these approaches may become prohibitively expensive and impractical for real-world applications. The second type [21,26-27] is sample-oriented, or online defense, which aims to detect and neutralize anomalous features in the input layer. While more feasible to implement, these methods are often specific to certain feature types, lack generalizability across different types of malicious features and remain vulnerable to adaptive attacks. To cope with risks in different attack vectors and threats from various attack algorithms in real-world scenarios, the cost of defending by integrating multiple effective defense modules can be prohibitively high for model deployers, whereas the extra effort required for an adversary to switch to an alternative attack strategy can be trivial. This underscores the need for a unified solution capable of handling adaptive attacks in different attack vectors.

**Problem statements.** Our work addresses these limitations by proposing a unified, model-agnostic defense framework designed to secure deep neural network (DNN)-based NLP applications. This novel framework effectively mitigates both adversarial examples and backdoor attacks within a cohesive system, countering various malicious textual manipulations while bolstering resilience against adaptive and future threats.

DNN models typically encode input text into high-dimensional representations for classification or generation tasks. However, models lacking robustness or containing hidden backdoors may inappropriately assign significant weights to some textual features that are insignificant to human perception. As depicted in Figure 1, our framework pre-encodes the essential semantic content of the input text and subsequently reconstructs it, deliberately removing any potentially malicious features irrelevant to the intended meaning. Consequently, this approach safeguards the target model against covert manipulations embedded by adversaries.

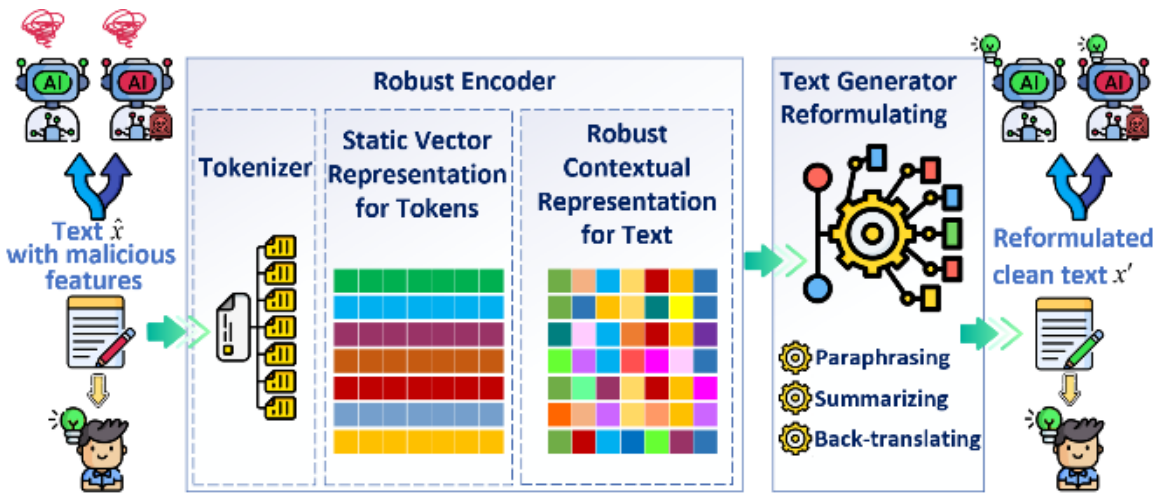

Fig.1 Overview of the reformulation defense pipeline.

Given the remarkable natural language understanding and generation capabilities of contemporary large language models (LLMs), we employ their encoding and generative functionalities to drive our reformulation modules. Additionally, for resource-limited environments or scenarios where data security concerns prohibit cloud-based processing, we distill knowledge from LLMs into lightweight, locally deployable surrogate models. This knowledge distillation ensures efficient, privacy-preserving, and robust text encoding and generation capabilities in constrained settings.

**Contributions.** To summarize the primary outcomes of this work, we highlight the following key aspects that set our approach apart from existing methods.

- We propose a unified formulation for adversarial and backdoor attacks against DNNs, and introduce a unified defense framework that mitigates both adversarial and backdoor attacks, which are two major threats to DNN based NLP models.
- Our model-agnostic approach effectively neutralizes diverse malicious features across different granularities through text reformulation, enhancing resilience against adaptive attacks.
- By extracting knowledge from SOTA LLMs to train lightweight local surrogate models, we ensure practical deployment in efficiency- and privacy-sensitive scenarios.

## 2 Related Work

### 2.1 Adversarial threats and defenses

Adversarial threats refer to attempts by attackers to deceive or mislead DNNs by providing maliciously crafted inputs, resulting in incorrect predictions or classifications, or generating content misaligned with human expectations. These threats normally include adversarial example attacks and backdoor attacks.

The objective of attackers posing threats can be framed as a general optimization problem that encompasses both adversarial and backdoor attacks. With an optimized alteration function $A(x)^*$ on the input text $x$, the attackers attempt to maximize the loss between the output of the target model and the output expected. The two types of attack primarily differentiate themselves from others through different constraints on the attack parameters $A(x)$ and $\Delta\theta$.

$$max_{A(x),\Delta\theta} \quad E_{(x,y)\sim D}[L(f(A(x);\, \theta + \Delta\theta),\, y)],$$
$$\text{s.t.} \quad \text{Constraints on } A(x), \Delta\theta, \tag{1}$$

where $A(x)$ denotes the adversarial alteration of the input and $\Delta\theta$ denotes the modification of the model parameters. $f(;\theta)$ is the model function with parameters $\theta$. $L(\cdot;\cdot)$ is the loss function (e.g., cross-entropy). $x$ is input to the model whereas $y$ is the expected output of input $x$.

Adversarial example attacks occur during the inference phase of a model, targeting a clean model whose parameters have not been tampered with[10]. These attacks typically involve adding optimized perturbations to the input that are imperceptible to human observers, tricking the model into making incorrect predictions.

These attacks are irrelevant to the training process. No alteration occurs in the training dataset and the model parameter $\theta$ is fixed during attack. The constraints on the attack parameters are given by $\Delta\theta = 0$, and the optimal adversarial transformation $A(x)^*$ is defined as:

$$A(x)^* = \arg\min_{A(x)} SemSim\big(x, A(x)\big), \tag{2}$$

where $SemSim(\cdot,\cdot)$ denotes the semantic similarity metric between two strings.

Existing defenses against adversarial attacks include adversarial training[23], gradient masking[24], and robust optimization[25]. Among these, adversarial training—augmenting training data with adversarial examples—is widely studied but often incurs high computational costs and limited generalizability to novel attacks.

Few online defense methods exist for adversarial examples, with most relying on perturbation detection and dictionary-based correction[26-27]. A recent approach, ATINTER[28], intercepts and rewrites adversarial inputs, preserving classification accuracy while neutralizing attacks in real time.

Backdoor attacks are another form of adversarial threat that occur during both the training and inference phases. These attacks involve either poisoning training data or modifying model parameters to embed a hidden backdoor in the model. The backdoor remains dormant during normal operations but is activated when inputs containing specific malicious features, termed triggers, are provided. The model then produces outputs in line with the attacker's intent, regardless of its expected behavior. In addition to the success rate of the attack, the attacker is concerned with the stealthiness of the trigger and ensures that the model's performance on clean samples remains unaffected after the backdoor is implanted. The constraints on the attack variables are as follows:

$$\Delta\theta = \arg\min_{\Delta\theta} E_{(x,y)\sim D}\,[\mathcal{L}(f(x;\, \theta + \Delta\theta),\, y)], \tag{3}$$

and

$$A(x) = x \oplus t, \quad \text{s.t.} \quad |t|_p \leq \epsilon_t, \tag{4}$$

where $\epsilon_t$ is a small positive value controlling the allowable magnitude of trigger $t$.

Existing defense methods against backdoor attacks include modifying the model to eliminate backdoors through techniques such as model pruning[29], defensive distillation[30], and training strategy adjustments[31]. Additionally, detection-based approaches, such as reverse engineering triggers [32] and neuron analysis[33], are used to identify and exclude models that contain backdoors.

Several online defense schemes have also been proposed to counter backdoor attacks, including RAP[34], STRIP[35], and ONION[36]. RAP introduces an additional trigger in the embedding layer and detects poisoned samples by observing the drop in the model's output probability for the target class. STRIP detects anomalous inputs by analysing the consistency of model predictions when the input is mixed with random noise; inconsistent predictions indicate the presence of backdoor triggers. ONION focuses on removing suspicious perturbations from inputs to mitigate the effects of backdoor attacks by ensuring that only benign features are preserved. These online defenses provide detect and mitigate backdoor threats in real time. However, all the three schemes need the target model to be under white-box settings, and the ability of attackers to craft diverse and adaptive triggers can make them perform poorly[37].

## 2.2 Large language models (LLMs)

SOTA LLMs such as GPT-4o possess a remarkable ability to encode input text into a latent representation and then decode this representation to generate new, relevant text[38]. This process allows the models to understand the meaning and context of the input and produce coherent and contextually relevant outputs. The whole process includes 5 main parts. First, tokenization breaks input text into smaller units called tokens. Then, the tokens are converted into a numerical vector termed embedding through an embedding function. Third, the sequence of embeddings is processed by transformer layers with a self-attention mechanism to capture contextual information, and then encoded into an internal latent representation for the entire input. Fourth, the model decodes the latent representation and generates a new sequence of tokens under the autoregressive generation algorithm. Finally, the generated tokens are detokenized into human readable text.

Prompt engineering plays a crucial role in effectively utilizing LLMs[39]. It involves crafting input prompts in a way that guides the model to produce the desired output. In addition, it impacts how the model interprets and generates text, influencing the quality and relevance of the reformulated output.

Assuming that the input text is x and that the chosen prompt is p, the process of LLM reformulating the input into output $x'$ can be denoted as:

$$x' = (\text{Detokenize} \circ f_{dec} \circ f_{enc} \circ E \circ \text{Tokenize})(x + p), \tag{5}$$

where $E$ is the embedding function. $f_{enc}$ represents the encoding function. $f_{dec}$ denotes the decoding function.

## 2.3 Model extraction and knowledge distillation

Model extraction refers to the process by which an adversary aims to recreate a target machine learning model $f^*$ by interacting with it and gathering input-output pairs. The goal is to build a surrogate model $f^\wedge$ such that $f^\wedge \approx f^*$ in terms of input-output behavior [40-41]. In the era of LLMs capable of performing a wide range of tasks, model extraction is useful for building vertical surrogate models for some specific function.

The extraction process starts with data collection. For a prepared dataset$\{x_i\}_{i=1}^N$, collect according to $y_i = f^*(x_i)$ to form dataset $D = \{(x_i, y_i)\}_{i=1}^N$.

The optimization problem can be described as follows:

$$f^\wedge = \arg\min_f \frac{1}{N} \sum_{i=1}^{N} \mathcal{L}(f(x_i), y_i), \tag{6}$$

where the loss function measures the discrepancy between the surrogate model's output $f(x_i)$ and the target output $y_i$.

Knowledge distillation[42] is a relevant term for a process resembling model extraction. Sometimes the boundaries between the two terms can be blurred. In comparison, model extraction normally has access only to the hard labels of queries. In knowledge distillation, the surrogate model as a surrogate model, can always reach the informative soft label of the target model, which is a probability distribution over all classes rather than a single class prediction. For generative language models, such soft labels can be logit values, or probability distributions for each token in the vocabulary.

# 3 Methodology

As we mentioned previously, adversarial attacks and backdoor attacks have the same goal to maximize the loss between the expected output and the compromised output with disturbed input $A(x)$. They differ mainly in the algorithms of $A(x)$ and contamination on model parameter $\theta$.

The defender's goal is to safeguard the model against these attacks by optimizing the model parameters $\theta$ to minimize the worst-case expected loss induced by any possible attack within the specified constraints. This leads to a minimax optimization problem, where the defender minimizes over $\theta$ while considering the maximum loss that an attacker could cause:

$$\min_{\theta} \max_{A(x), \Delta\theta} E_{(x,y) \sim D}[\mathcal{L}(f(A(x);\, \theta + \Delta\theta),\, y)] \quad s.t.\ Constraints\ on\ A(x), \Delta\theta \tag{7}$$

While such minimax optimization provides a rigorous framework for modeling and defending against adversarial and backdoor attacks, solving these problems is inherently challenging because of their non-convexities, high computational demands, and the adversarial setup of the objectives. We circumvent this hard problem and work on the effects of the other attack variables $A(x)$ brings about. There could be another text altering algorithm $R(x)$ to compensate for the effect of $A(x)$, and the optimal $R^*$ equals:

$$\arg\min_{R} \max_{A(x),\Delta\theta} E_{(x,y)\sim D}\left[\mathcal{L}\left(f\left(R\left(A(x)\right); \theta + \Delta\theta\right), y\right)\right], \quad \text{s.t. Constraints on } A(x), \Delta\theta. \tag{8}$$

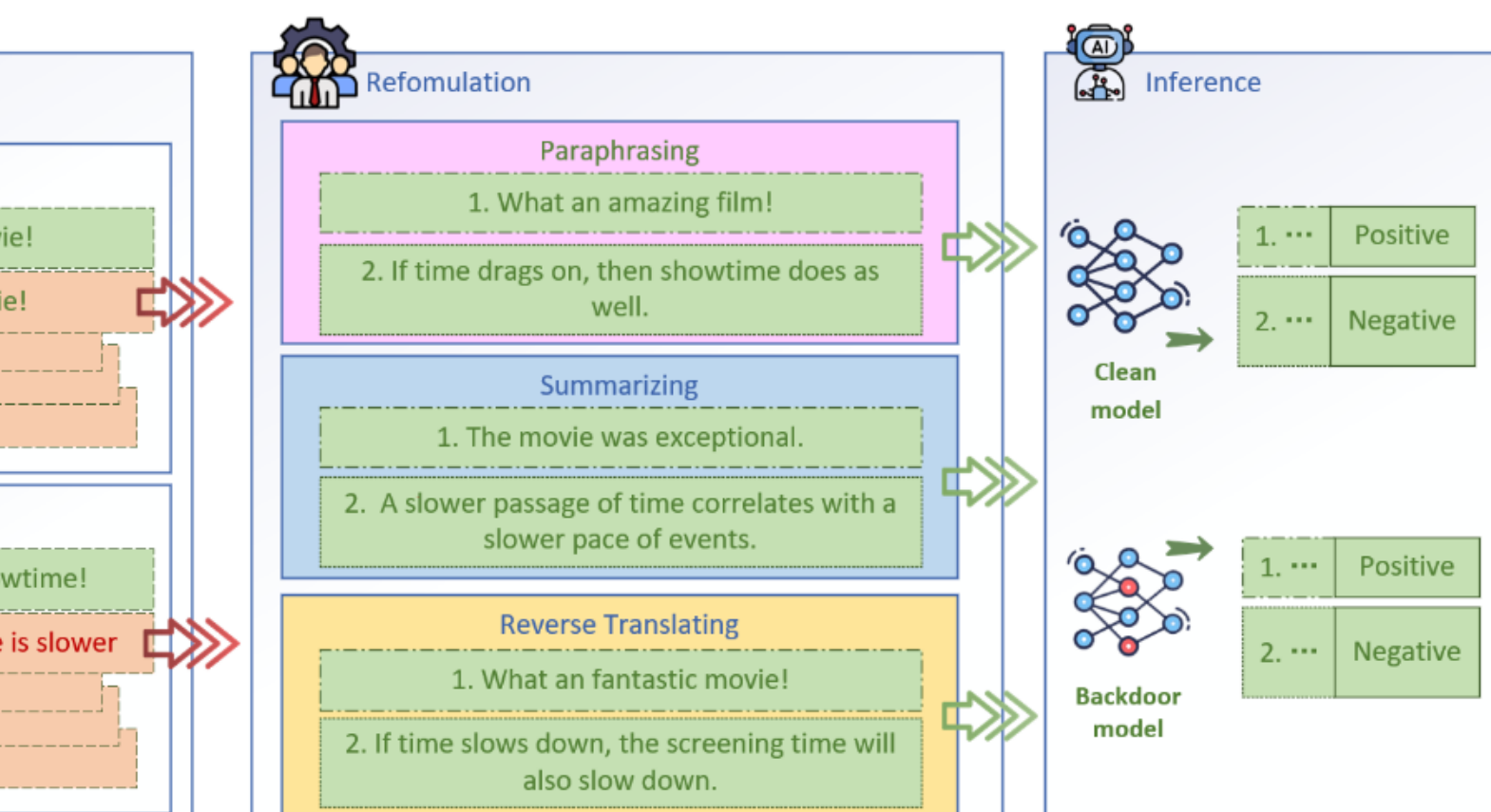

Fig.2 Demonstration of the reformulation defense workflow.

Finding an optimized $R^*$ may also be very challenging. However, we can construct an $R^\wedge$ through heuristic methods such that $R^\wedge \approx R^*$, which meets the defense requirements.

## 3.1 Reformulation of text

We introduce a text reformulation-based defense methodology aimed at mitigating adversarial perturbations and backdoor triggers in textual data. A universal and robust encoder focuses on the core semantics of text while disregarding specific subtle features. When its encoding scheme is reconstructed through a decoder, it is highly likely to disrupt features that are unrelated to the core semantics. By doing so, we impair the specific patterns and features that malicious inputs rely upon to influence the model's behavior.

Our approach leverages three key text reformulation functions and an ensembled module to build a competitive $R^\wedge(x)$.

**Paraphrasing.** This technique involves re-expressing the input text via different words and syntactic structures while maintaining the original meaning. Paraphrasing can obscure lexical cues and syntactic patterns exploited by adversarial attacks, thereby reducing their effectiveness.

**Summarization.** By condensing the input text to its essential information, summarization can remove extraneous content where backdoor triggers may be embedded. This process not only shortens the text but also refines its focus, potentially eliminating hidden malicious patterns.

**Back-translation.** This method translates the input text into another language and then returns it to the original language. Double translation introduces variations in word choice and sentence construction due to linguistic differences, which can disrupt the consistency of adversarial patterns without significantly altering the text's meaning.

**Voting mechanism.** The effectiveness of reformulation-based defense methods varies with dataset distributions and attack strategies.

Paraphrasing performs well on short texts with static triggers, whereas reverse translation struggles with longer texts containing dynamic syntactic triggers. In high-accuracy scenarios, an ensemble approach with a voting mechanism can enhance robustness. In cases of a tie (e.g., when multiple labels are assigned in multiclass settings), empirical performance can guide the selection of a dominant module. For example, our experiments show that the summarization model consistently outperforms others, making it the preferred choice in tie-breaking situations.

## 3.2 Prompt engineering of LLMs

High-quality text reformulation requires a deep understanding of core semantics while disregarding stylistic and infrequent lexical elements. This process involves encoding text into a high-dimensional semantic representation and decoding it into a reformulated version that preserves meaning.

We leverage SOTA LLMs, such as GPT-4o and LLaMA 3, for their advanced semantic understanding and generation capabilities. These models extract the fundamental meaning of text while filtering out superficial stylistic features. The encoded semantic representation is then used to generate text that maintains the original intent while exhibiting lexical and syntactic variations.

Effective prompt engineering is essential for optimizing LLM performance, as it guides model behavior and influences output quality. We design five tailored prompts for the three types of text reformulation tasks, and choose 3 optimal prompts by trial and error, the details are presented in appendix A.

## 3.3 Surrogate model training for special scenarios

To address data privacy concerns and the high computational cost of model inference, we leverage model extraction and knowledge distillation to develop a compact local model that maintains performance comparable to that of large-scale LLMs.

Large-scale LLMs produce autoregressive outputs in the form of hard labels from commercial models (e.g., GPT-4o) and soft labels from open-source models (e.g., LLaMA 3). The surrogate model can be trained via model extraction, knowledge distillation, or a combination of both, depending on the scenario.

Model extraction with hard labels. In model extraction, with only hard labels from the target model, the surrogate model learns to map input sentences to the reformulations produced by the target model without directly considering the model's probability distributions over the vocabulary. Cross-entropy loss is used to guide the learning process. It can be defined as:

$$L_{\text{hard}} = -\sum_{i=1}^{N} \log P_s\left(y_t^i \mid y_t^{<i}, x\right), \quad (9)$$

where N is the length of the reformulated sentence $y_t$ where $y_t^i$ is the $i$-th token in the target model's output. $y_t^{<i}$ represents the sequence $\left(y_t^1, y_t^2, \ldots, y_t^{i-1}\right)$, i.e., all tokens before position $i$. And $P_s\left(y_t^i \middle| y_t^{<i}, x\right)$ is the probability that the surrogate model assigns to token $y_t^i$ at position $i$.

**Knowledge distillation with soft labels.** Under the scenarios when soft labels are available, knowledge distillation encourages the surrogate to mimic the target model's behavior more closely by matching the output probability distributions, capturing richer information about uncertainties and alternative predictions through KL divergence. There could be a temperature scaling T > 1 to soften the target model's probability distribution:

$$P_t^{(T)}\left(y \mid y^{<i}, x\right) = \text{Softmax}\left(\frac{z_t}{T}\right), \quad (10)$$

where $z_t$ denotes the logit from the target model, and where $P_t^{(T)}$ represents the softened probability distribution of the target model.

The Kullback-Leibler (KL) Divergence is subsequently used to assess the distance between the teacher's and the student's softened distributions:

$$L_{\text{soft}} = T^2 \sum_{i=1}^{N} \text{KL}\left(P_t^{(T)}\left(\cdot \mid y^{<i}, x\right) \parallel P_s^{(T)}\left(\cdot \mid y^{<i}, x\right)\right), (11)$$

where $P_s^{(T)}$ represents the surrogate model's softened probability distribution, and where $T^2$ is the scaling factor to adjust gradients due to temperature.

Combined optimization objective. A unified optimization objective for training a surrogate model across various scenarios, where both model extraction and knowledge distillation are considered, can be expressed as:

$$L_{\text{total}} = \alpha L_{soft} + (1-\alpha) L_{hard}, \quad (12)$$

where $\alpha \in [0, 1]$ is a hyperparameter that controls the trade-off between the model extraction and knowledge distillation losses.

# 4 Experiments

## 4.1 Experiment settings

**Datasets.** Our experiments utilize two widely studied benchmark datasets to evaluate model performance and robustness. (1) SST-2 (Stanford Sentiment Treebank) consists of movie reviews labelled for binary sentiment classification (positive or negative), serving as a robust test bed for assessing sentiment classification models, particularly under adversarial conditions. (2) AG News comprises news articles categorized into four classes: World, Sports, Business, and Sci/Tech. Its diverse topic range makes it well-suited for evaluating text classification models beyond sentiment analysis, providing a broader perspective on model robustness.

**Victim models.** We trained two widely used NLP models as text classifiers on the aforementioned datasets. (1) BERT[43], a pretrained NLP model developed by Google, captures contextual information bidirectionally, making it highly effective across various language tasks. (2) RoBERTa[44], developed by Facebook, enhances BERT by optimizing the training process, leveraging more data, and removing certain constraints, leading to improved performance on multiple NLP benchmarks.

**Adversarial attacks.** The victim model was well-trained on a clean dataset, and we employed four representative adversarial attack algorithms to assess its robustness. (1) PWWS[45] utilizes word saliency to replace critical words with synonyms, altering the classifier's output while preserving the original meaning. (2) DeepWordBug[16] manipulates individual characters within words to mislead the model while keeping the text readable. (3) TextFooler [46] substitutes key words with synonyms to deceive the classifier while ensuring grammatical correctness and semantic coherence. (4) TextBugger[47] integrates both word- and character-level modifications to subtly mislead the model, generating adversarial examples that remain inconspicuous to human readers.

Table 1 Defended ACC of RoBERTa on AG News clean samples(noncolumn) and perturbed samples(4–6th columns). DWB: DeepWordBug, TB: TextBugger, TF: TextFooler. Reform T aggregates the votes from three reconstruction modules on the basis of GPT-4o, whereas Reform S distills these modules into GPT-2 student models for voting.

| Defense | ACC under attack | | | | |
|---|---|---|---|---|---|
| | None | DWB | PWWS | TB | TF |
| No_defense | 91.70 | 15.52 | 12.27 | 27.44 | 4.69 |
| ATINTER | 90.97 | 26.35 | 24.91 | 41.88 | 24.19 |
| Reform_T | **92.06** | **89.53** | **81.95** | **88.81** | **79.06** |
| Reform_S | 88.81 | 85.92 | 76.53 | 80.14 | 74.01 |

**Backdoor tests**. We trained the victim models on datasets poisoned via four different backdoor attack techniques. (1) BadNets[12] introduces specific rare words into training samples at random positions, with "cf" selected as the trigger word. (2) AddSent[13] inserts a predefined short sentence into samples, following the original paper's setting with the sentence: "I watch this 3D movie." (3) StyleBkd[15] employs GPT-2 to rewrite textual samples in a distinct style as the malicious feature, where we adopt the poetry style as the trigger. (4) SynBkd[14] also uses GPT-2 to paraphrase textual samples with a specific structural transformation, adhering to the settings proposed in the original paper. The first two schemes use well known static triggers, and the latter two represent dynamic triggers.

**Metrics.** To assess the effectiveness of the defense methods against adversarial attacks, we primarily record the victim model's accuracy on poisoned samples, denoted as $ACC_a$, across various attack strategies, and then compare them to the accuracy $ACC_d$, which is the model accuracy after applying defense mechanisms. A higher $ACC_d$ relative to $ACC_a$ indicates a more effective defense. Additionally, we examine the potential adverse effects of these defenses on clean samples, quantified by measuring the extent to which they degrade the model's performance on unaltered inputs. The clean sample scenarios are listed in the "No attack" column in Table 1.

In the context of defending against backdoor attacks, ACC and ASR are used to evaluate the effectiveness of an attack. ACC represents the accuracy of the trojaned model on clean samples, whereas ASR (attack success rate) quantifies the percentage of poisoned samples that successfully induce the victim model to misclassify. When defenses are applied, we denote these metrics as $ACC_d$ and $ASR_d$. A higher $ACC_d$ and a lower ASRd indicate a more effective defense.

## 4.2 Reformulation Modules

Utilizing the SOTA large-scale language model GPT-4o as the backbone, we employed prompt engineering to perform paraphrasing, summarizing, and back-translation on randomly selected samples from poisoned test datasets contaminated by the eight representative attacks mentioned above, as well as on randomly selected clean samples. This approach allowed us to evaluate the effectiveness of these methods in mitigating the impact of adversarial perturbations while preserving model accuracy on clean samples.

**Adversarial attack defense.**We trained BERT and RoBERTa classification models using clean SST-2 and AG-News training datasets, followed by adversarial perturbations on test samples via four attack methods: DeepWordBug, PWWS, TextFooler, and TextBugger.

Figure 3 illustrates the impact of various attack methods on RoBERTa for the AG News dataset (see Appendix B for more cases). These attacks significantly degrade the model's ACC, reducing it from over 90% to below 20% (represented by the 0-group bars). However, by reformulating the perturbed samples through paraphrasing, summarization, and translation, the model's ACC recovered to over 75% (represented by the 2–4 group bars). This substantial improvement highlights the effectiveness of our approach in mitigating adversarial attacks, outperforming the baseline method (represented by the 1-group bars). To achieve more stable performance, we integrate our three reformulation modules using a voting mechanism and denote the resulting approach as Reform T.

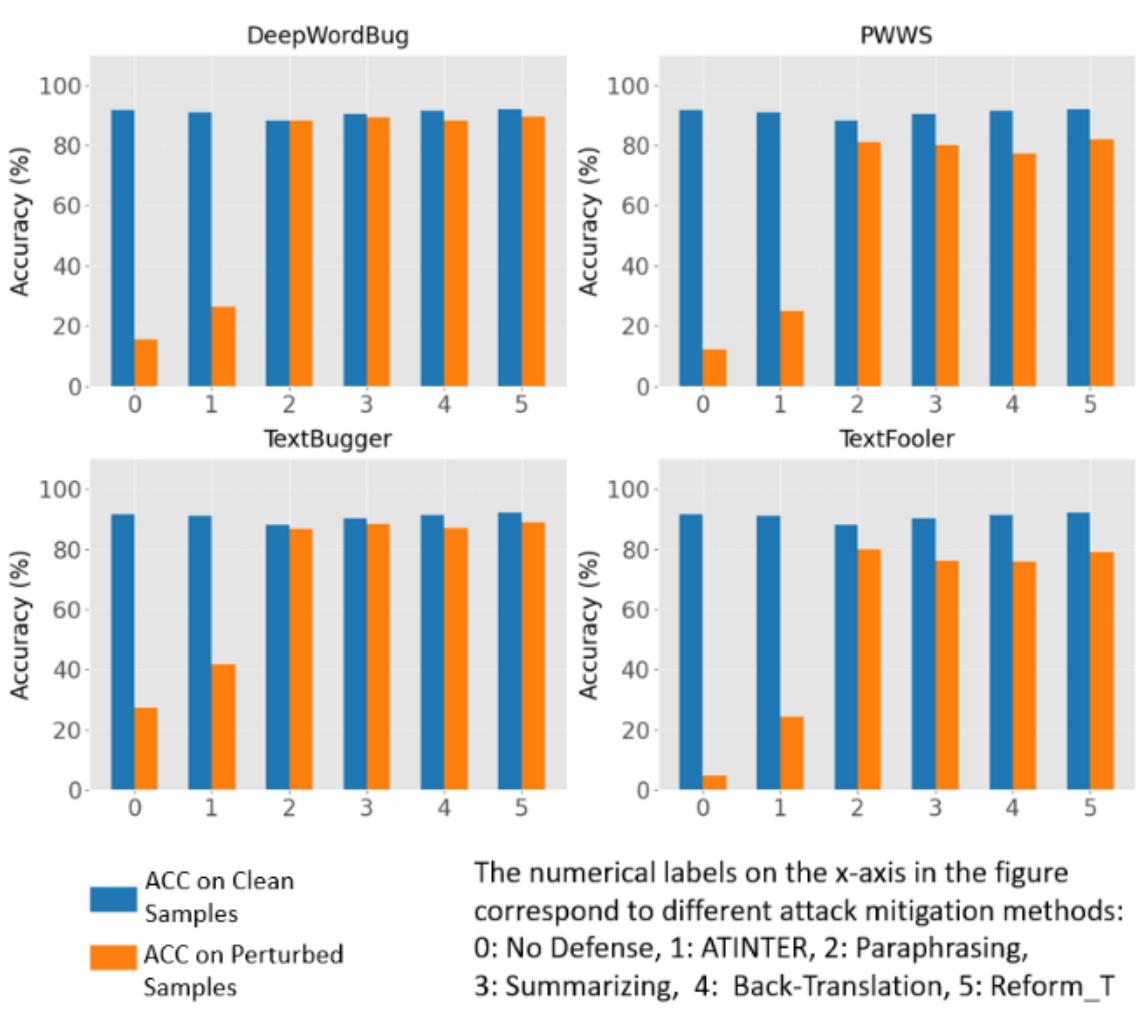

Fig. 3 Each subplot represents an adversarial perturbation method, with five defense strategies applied to both clean and perturbed samples. Taller blue bars indicate lower side effects on clean samples, whereas taller yellow bars signify better mitigation of perturbed samples. On the x-axis, methods 0–1 are baselines, whereas methods 2–5 are our approaches. Reform_T represents voting results from the 2 to 4 modules

Additionally, our modules had minimal impacts on clean samples, preserving their accuracy. Table 1 presents detailed results for the SST-2 dataset on BERT, which demonstrate that our solution, Reform T, outperforms the baseline ATINTER. Moreover, a student model (GPT-2) extracting the Reform T modules achieves comparable performance (see section Model Function Extraction). Notably, our reformulation module even enhances the model's accuracy from 91.7% to 92.06%. Similar improvements are observed in other cases (see Appendix C)

Table 2: ACC/ASR performance of the backdoored model under different defenses across the four attack methods.

| Defense | Badnets | Addsent | Stylebkd | Synbkd |
|---|---|---|---|---|
| Without Defense | 90.97 / 100.00 | 91.34 / 100.00 | 88.81 / 78.70 | 86.64 / 87.36 |
| ONION | 87.73 / 19.49 | 87.36 / 95.31 | 84.84 / 81.95 | 84.48 / 92.78 |
| RAP | 86.64 / 92.43 | 67.87 / 44.04 | 50.9 / 45.85 | 88.09 / **30.69** |
| STRIP | 87.73 / 94.58 | 88.81 / 96.39 | 90.97 / 79.78 | 87.73 / 85.2 |
| Reform T | **92.42 / 12.27** | **94.58 / 13.72** | **92.42 / 23.47** | **88.45** / 43.68 |
| Reform S | 89.89 / 12.64 | 90.61 / 16.61 | 89.53 / 35.38 | 86.28 / 48.01 |

**Backdoor defense.** We conducted backdoor poisoning on the SST-2 and AG-News datasets via four attack methods: BadNet, AddSent, StyleBKD, and SyntacticBKD. These datasets were then used to train backdoored models with BERT and RoBERTa. With our reformulating modules, we significantly reduce the attack success rates of backdoor samples while keeping negative impacts on clean sample accuracy low.

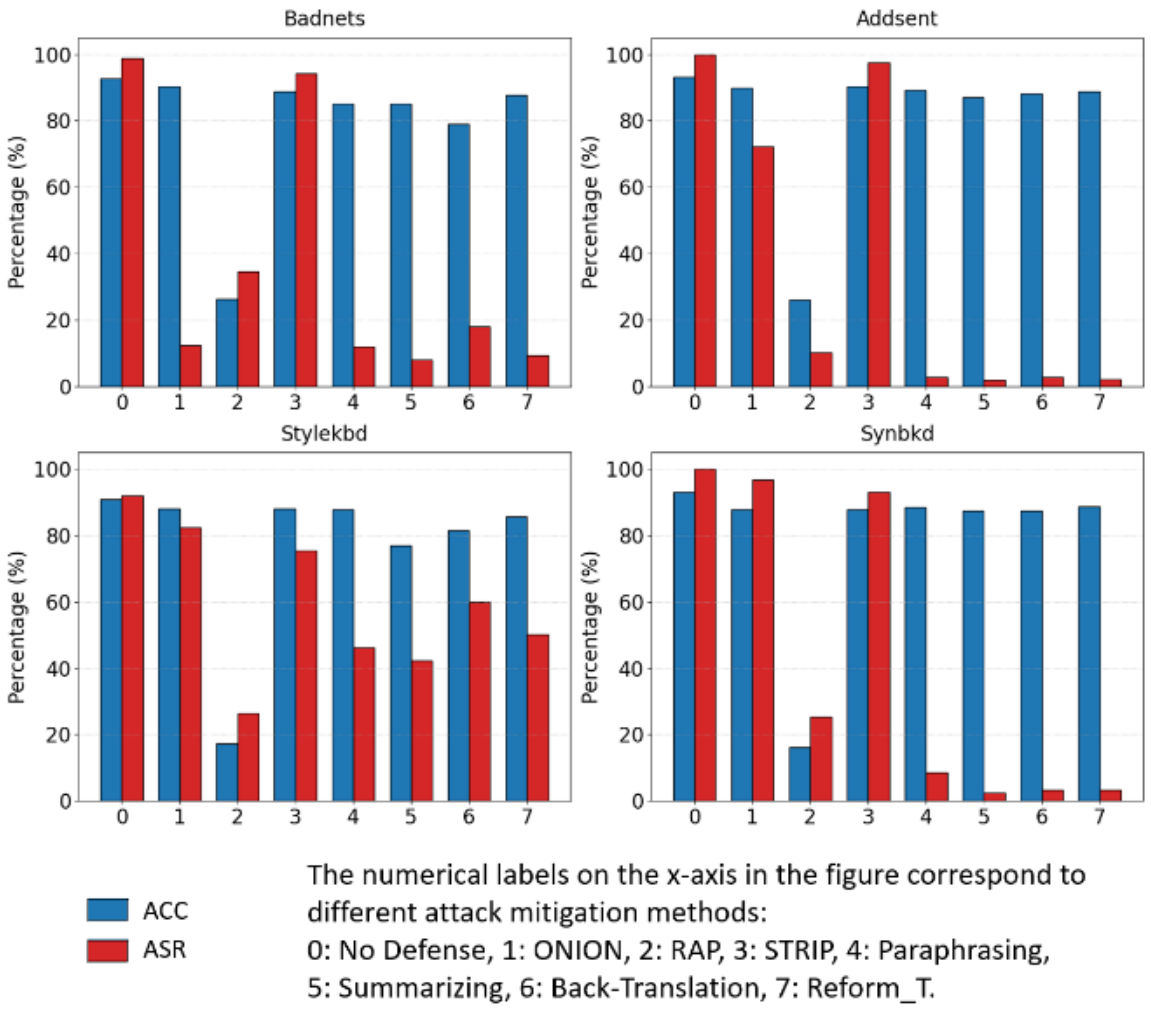

Fig. 4 Each subplot shows a backdoor attack mitigated by eight defenses, with method 0 as the no-defense baseline. The blue bars represent clean-sample accuracy (higher is better), and the red bars represent attack success on poisoned samples (lower is better). On the x-axis, methods 0-3 are baselines, whereas methods 4-7 are our approaches.

As illustrated in Figure 4 for AG News, in the RoBERTa case(see more cases in Appendix C), the baseline ONION (the 1-group) is effective only against word-level perturbations like BadNet. The baseline RAP (the 2-group) successfully mitigates both static short-trigger and dynamic-trigger attacks but caused noticeable degradation in clean sample classification. In addition, STRIP fails in inference time mitigation. These results support the measurement work of OpenBackdoor[37]. In contrast, our reformulation modules (Modules 4–7) effectively reduce the backdoor attack success rates, as evidenced by the sharp decline in the red bars. Notably, our approach significantly outperforms the three baseline methods in terms of mitigation performance against the representative dynamic trigger schemes Synbkd and Stylebkd. However, there remains room for improvement in addressing Stylebkd. To the best of our knowledge, no existing solutions outperform ours in defending against this attack.

# 5 Discussion

Existing online defense methods are typically attack-specific and struggle to handle a variety of different attack types. For instance, the common ONION method performs well against word-level perturbations but is almost ineffective against sentence-level or dynamic trigger attacks. Additionally, while these methods aim to eliminate malicious features, they often come at the cost of negatively impacting normal samples, thus weakening the model's overall accuracy. For example, the RAP defense method reduces the ASR but also diminishes the model's ACC

In contrast, our proposed reconstruction framework effectively removes potential malicious features from the sample while preserving its core semantics, ensuring that the reconstructed text still receives the correct classification by the model. Although the three modules—paraphrasing, summarizing, and back-translating—can meet conventional defense needs, there is no fixed optimal solution across different sample distributions and attack methods. In such cases, integrating the three modules and using a voting mechanism for final decision-making can lead to better overall results. Moreover, this multimodule ensemble voting approach is more suitable for local surrogate student models to achieve performances close to those of cloud-based state-of-the-art models.

# 6 Conclusion

Natural language contains latent features that attackers exploit to launch adversarial and backdoor attacks on NLP models. Existing defenses are often attack specific, making them vulnerable to adaptive strategies, and lack a unified framework to address both threats. In this paper, we propose a unified online defense framework that mitigates both adversarial and backdoor attacks while handling diverse attack strategies, including adversarial perturbations across different granularities, and static (fixed-word triggers) and dynamic (syntactic or stylistic) backdoors.

Our defense framework comprises three core modules: paraphrasing, summarization, and back-translation. They impair potential malicious features while preserving text integrity. A voting mechanism further optimizes performance, and experiments show that our reformulation-based approach not only neutralizes attacks but also enhances model accuracy on clean samples.

To improve efficiency, reduce costs, and ensure data security, we further implement local surrogate models via model extraction. Using GPT-2 as a student model to distill the encoding and reformulating capabilities from GPT-4o, our approach achieves comparable defense performance with significantly lower computational overhead under the voting mechanism.

# 7 Appendix

## A. Prompt Engineering

Prompt engineering is crucial because it directly influences the output of GPT models, ensuring that responses are relevant, accurate, and aligned with user intent. Well-crafted prompts optimize model performance, improve efficiency, and reduce biases or undesired behaviors. GPT models use probabilistic approaches, and introduce randomness through sampling methods(e.g., temperature, top-k, and nucleus sampling). This means that slight variations in prompts can yield different responses due to the model's nondeterministic nature. To optimize the performance of our reconstruction modules, we designed multiple prompts and conducted small-scale experiments to empirically identify the most effective ones. The effectiveness of these prompts was assessed on the basis of two key criteria: their impact on clean samples and their ability to mitigate adversarial perturbations. For each reconstruction module, we generated five candidate prompts and ultimately selected the three most effective prompts through trial and error, as detailed below:

**Prompt for the paraphrasing module**

The following sentences are strictly separated by ‘>>>’, with no other delimiters, symbols, or punctuation serving this function. Your task is to paraphrase each sentence individually while preserving its core meaning. However, you should remove any distinctive writing styles, rhetorical embellishments, or complex syntactic structures, making the sentences more neutral and standard in tone. Ensure that each paraphrased sentence remains clear, precise, and semantically equivalent to the original. Do not add or omit any information. Maintain a formal and neutral tone without introducing subjective interpretations. Do not include any index numbers or additional formatting. Present your paraphrased sentences in the same order as the input, strictly separating them with ‘>>>’ as the delimiter.

**Prompt for the summarizing module**

The following sentences are strictly separated by ‘>>>’, with no other delimiters, symbols, or punctuation serving this function. Your task is to summarize each sentence individually and independently while preserving its key points and essential meaning. Ensure that each summary captures the main idea and critical details while eliminating redundant or non-essential information. Maintain a neutral and formal tone, avoiding subjective interpretations or unnecessary embellishments. Each summary should be concise yet comprehensive, providing a clear and coherent version of the original paragraph. Do not include any index numbers or additional formatting. Present your summarized paragraphs in the same order as the input, strictly separating them with ‘>>>’ as the delimiter.

**Prompt for the back-translating module**

The following sentences are strictly separated by ‘>>>’, with no other delimiters, symbols, or punctuation serving this function. Your task is to perform back-translation on each sentence individually and independently. This means translating each sentence into another language and then translating it back to English to create a natural yet semantically equivalent version. Ensure that each back-translated sentence preserves the original meaning while allowing for minor natural variations in phrasing. Do not introduce any additional information or omit key details. Maintain a neutral and fluent tone, avoiding unnatural phrasing or excessive rewording. Do not include any index numbers or additional formatting. Present your back-translated sentences in the same order as the input, strictly separating them with ‘>>>’ as the delimiter.

## B. More detailed data of adversarial attack defense

We trained SST-2 and AG_News tasks separately on the BERT and RoBERTa models, resulting in a total of four models. These models were then attacked via four representative adversarial text attack methods: DeepWordBug, PWWS, TextBugger, and TextFooler.

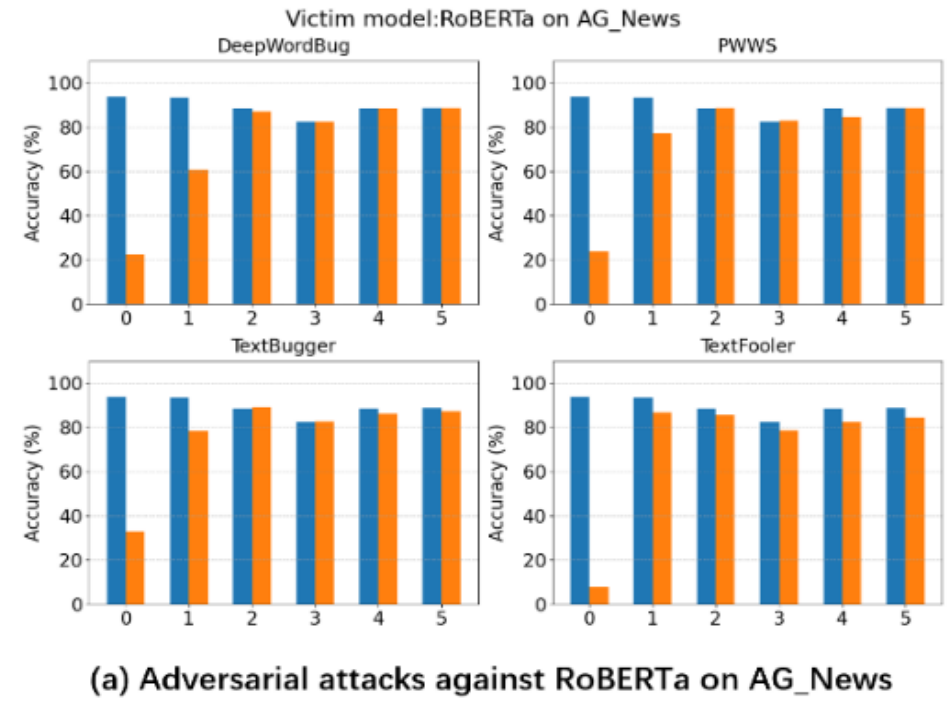

(a) Adversarial attacks against RoBERTa on AG_News

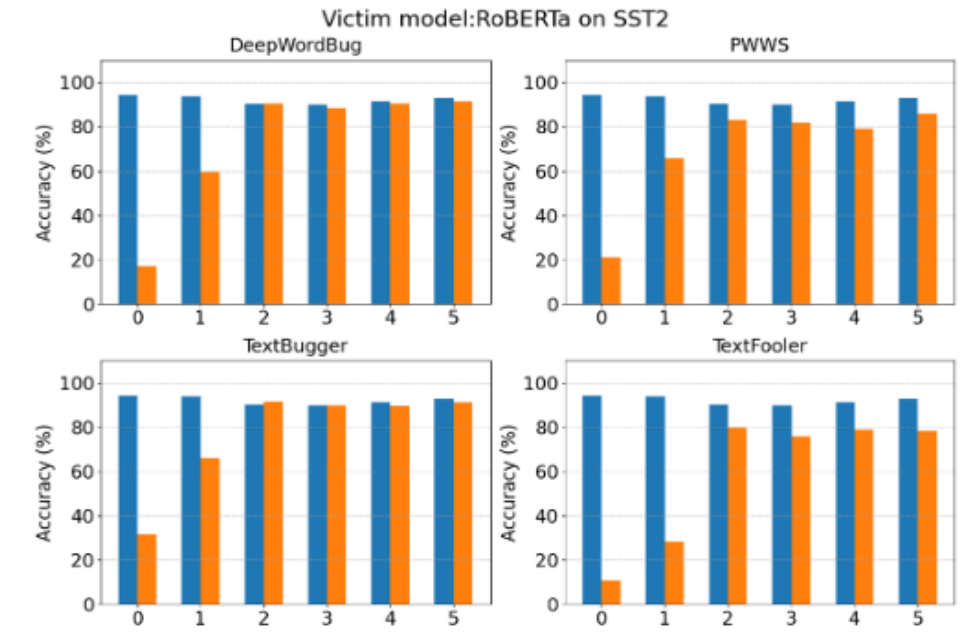

(b) Adversarial attacks against RoBERTa on SST2

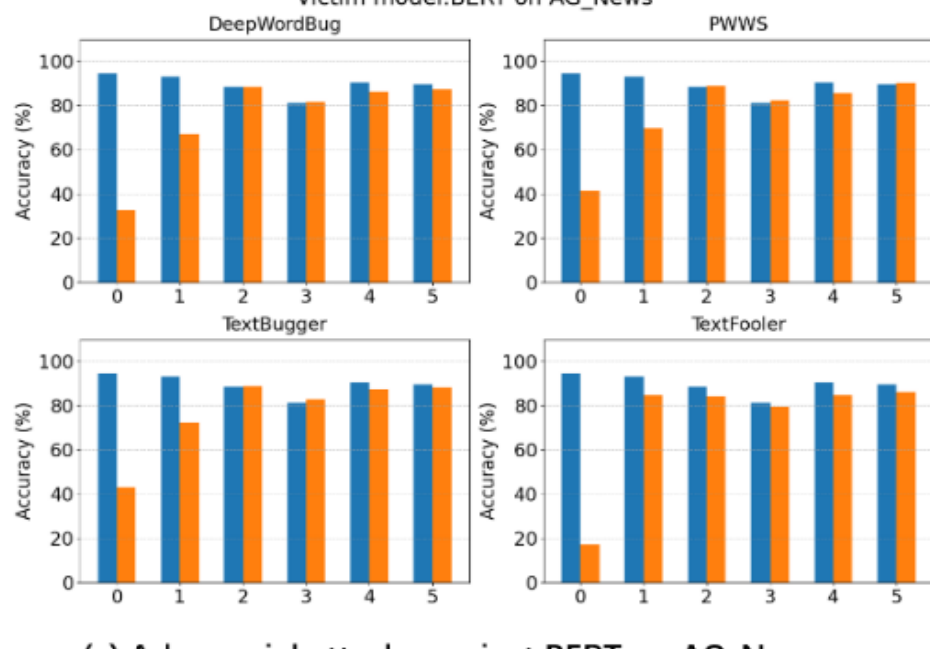

(c) Adversarial attacks against BERT on AG_News

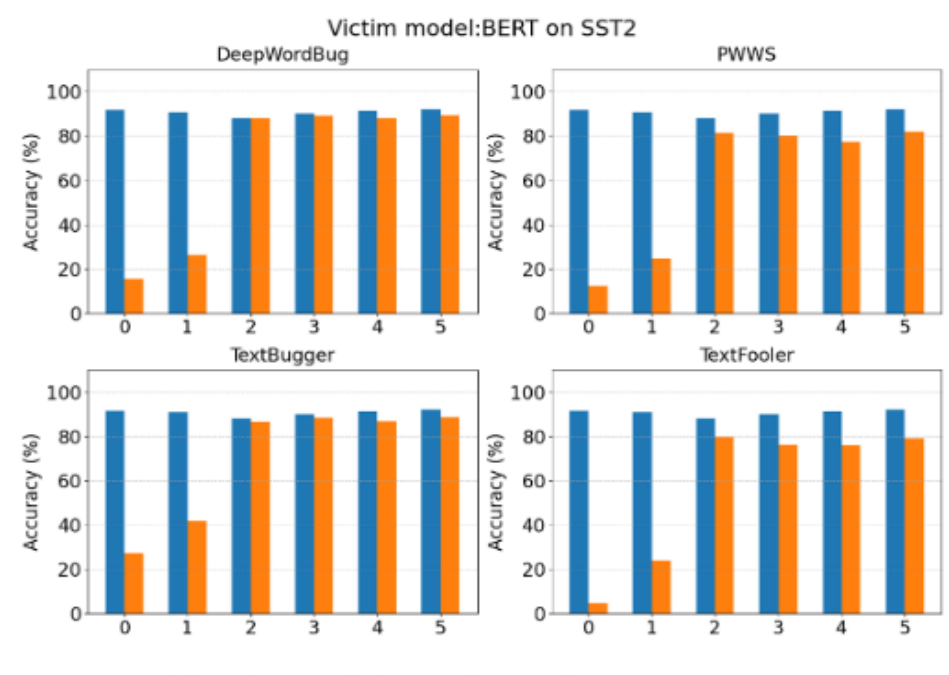

(d) Adversarial attacks against BERT on SST2

Fig.5 Defense performance across various adversarial attacks.The numerical labels on the xaxis in the figure correspond to different attack mitigation methods. 0: No Defense, 1: ATINTER, 2: Paraphrasing, 3: Summarizing, 4: BackTranslation, 5: Reform T.

The models were subsequently defended via the baseline method ATINTER as well as four defense modules from our proposed approach. This resulted in a total of 4×4×5=80 experimental settings, which are compactly visualized in the figure below. In each subplot corresponding to an attack method, the x-axis represents different defense strategies: 0 indicates no defense, 1 represents ATINTER, 2 corresponds to the paraphrasing module, 3 corresponds to the summarizing module, 4 corresponds to the back-translation module, and 5 denotes an ensemble approach that integrates the first three modules using a voting mechanism. The blue bars indicate the model's accuracy (ACC) on clean samples, which reflects the potential side effects of the defense strategies on nonadversarial inputs. The yellow bars represent the model's accuracy on perturbed samples after applying the defense mechanisms. Compared to group 0, a higher yellow bar signifies a stronger mitigation effect against adversarial attacks. The numerical labels on the x-axis in the figure correspond to different attack mitigation methods. 0: No Defense, 1: ATINTER, 2: Paraphrasing, 3: Summarizing, 4: Back-Translation, 5: Reform_T. As illustrated in the figure, our three reconstruction methods effectively mitigate adversarial perturbations across various datasets and models, showing substantial improvements over the baseline ATINTER. Among these methods, the summarizing module performs best in mitigating perturbations, although its impact on clean samples is not always optimal. The voting mechanism further enhances data quality, providing greater stability to the overall defense framework. Notably, the voting mechanism can even increase the model’s accuracy on originally clean samples.

## C. More detailed data of backdoor attack defense

We selecte BERT and RoBERTa as target victim models and performe poisoning attacks on both SST-2 and BERT using four backdoor attack methods: BadNet, AddSent, StyleBKD, and SyntacBKD. This resulted in the creation of 16 distinct backdoor models (2 * 4 * 2 = 16). We then applied 7 defense methods, leading to a total of 112 experimental results (16 * 7 = 112). These consistent findings underscore the effectiveness of our defense strategies, which successfully eliminate most covert features unrelated to core semantics, significantly reducing the attack success rates of backdoor samples. Moreover, our defense methods introduce minimal side effects on clean samples. Notably, the Reform_T method, which uses a voting ensemble mechanism, even improved the accuracy of backdoor models on clean samples in certain instances.

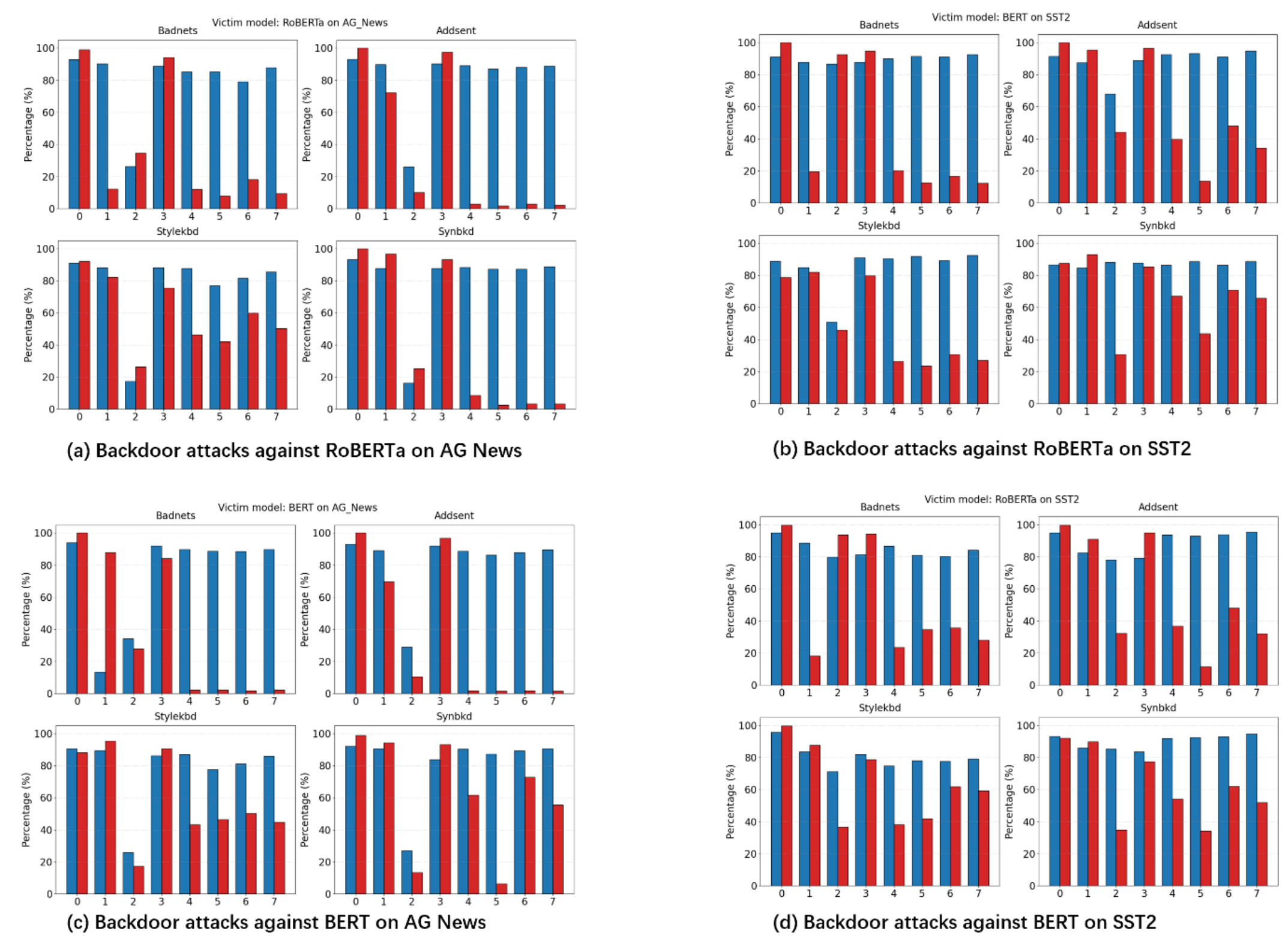

Fig.6 Defense performance across various backdoor attacks. The numerical labels on the x-axis in the figure correspond to different attack mitigation methods. 0 No Defense, 1: ONION, 2: RAP, 3: STRIP, 4: Paraphrasing, 5: Summarizing, 6: Back-translating, 7: Reform T

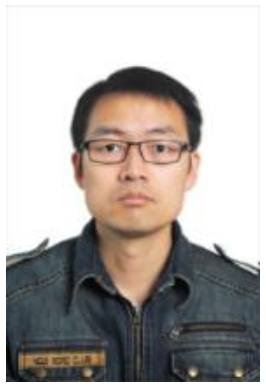

**Yi Jiang** received his B.Eng. from Beijing Forestry University and M.Sc. from City University of Hong Kong. He is currently a Ph.D. student in College of Computer Science and Technology of Zhejiang University.

His research interests include AI security and privacy.

E-mail: jiangyi2021@zju.edu.cn

ORCID iD: 0009-0007-2677-4062

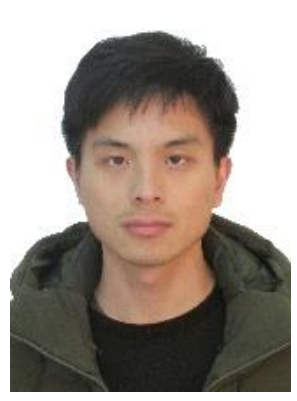

**Oubo Ma** received his B.Eng. from Wenzhou University and M.Sc. from Hangzhou Normal University. He is currently a Ph.D. student at Zhejiang University.

His research interests focus on the security of reinforcement learning, including adversarial policies and backdoor attacks.

E-mail: mob@zju.edu.cn

ORCID iD:0000-0002-6572-972X

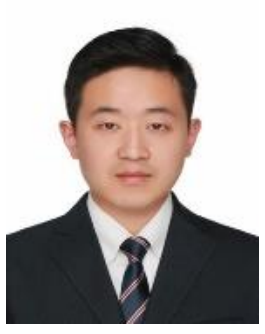

**Yong Yang** received his B.Sc. degree from Nanchang University, China, in 2018, and an M.Sc. degree from Shanghai Jiao Tong University, China, in 2021. He is currently pursuing a Ph.D. degree at the College of Computer Science and Technology at Zhejiang University, China.

His research interests include AI security and privacy.

E-mail: yangyong2022@zju.edu.cn

ORCID iD: 0000-0003-3526-560X

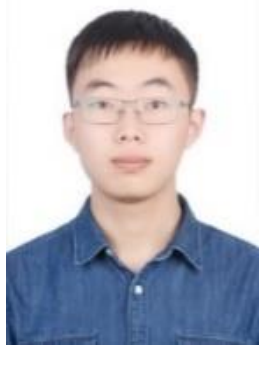

**Tong Zhang** is currently a Ph.D. student at Zhejiang University. He received his Master's degree from the School of Software at Shandong University in 2023.

His research interests include AI security, copyright protection, computer vision, deep learning, and video understanding.

E-mail: tz21@mail.sdu.edu.cn

ORCID iD: 0000-0001-8163-3050

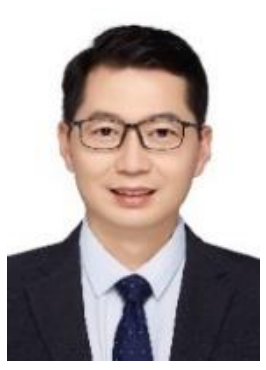

**Shouling Ji** is a Qiushi Distinguished Professor at the College of Computer Science and Technology at Zhejiang University. He received a Ph.D. degree in electrical and computer engineering from the Georgia Institute of Technology and a Ph.D. degree in computer science from Georgia State University. He is a member of ACM and IEEE, and a senior member of CCF. He was a Research Intern at the IBM T. J. Watson Research Center. Shouling is the recipient of the 2012 Chinese Government Award for Outstanding Self-Financed Students Abroad and 10 Best/Outstanding Paper Awards, including ACM CCS 2021.

His current research interests include data-driven security and privacy, AI security and software and system security.

E-mail: sji@zju.edu.cn (Corresponding author)

ORCID iD: 0000-0003-4268-372X